# Physics-informed UNets for Discovering Hidden Elasticity in Heterogeneous Materials

Ali Kamali[a], Kaveh Laksari[a,b,†]

a – Department of Biomedical Engineering, University of Arizona College of Engineering, Tucson, AZ

b – Department of Aerospace and Mechanical Engineering, University of Arizona College of Engineering, Tucson, AZ

[†] Corresponding author, Email Address: klaksari@arizona.edu, Phone number: +1 520 621 8124, Postal Address: 335, Bioscience Research Laboratories, 1230 N Cherry Ave, Tucson, AZ, 85721

# Abstract

Soft biological tissues often have complex mechanical properties due to variation in structural components. In this paper, we develop a novel UNet-based neural network model for inversion in elasticity (El-UNet) to infer the spatial distributions of mechanical parameters from strain maps as input images, normal stress boundary conditions, and domain physics information. We show superior performance – both in terms of accuracy and computational cost – by El-UNet compared to fully-connected physics-informed neural networks in estimating unknown parameters and stress distributions for isotropic linear elasticity. We characterize different variations of El-UNet and propose a self-adaptive spatial loss weighting approach. To validate our inversion models, we performed various finite-element simulations of isotropic domains with heterogenous distributions of material parameters to generate synthetic data. El-UNet is faster and more accurate than the fully-connected physics-informed implementation in resolving the distribution of unknown fields. Among the tested models, the self-adaptive spatially weighted models had the most accurate reconstructions in equal computation times. The learned spatial weighting distribution visibly corresponded to regions that the unweighted models were resolving inaccurately. Our work demonstrates a computationally efficient inversion algorithm for elasticity imaging using convolutional neural networks and presents a potential fast framework for three-dimensional inverse elasticity problems that have proven unachievable through previously proposed methods.

# 1 Introduction

Elasticity imaging is a technique to reconstruct the spatial distribution of mechanical properties using available deformation and force measurements. The mathematical problem in quasi-static elasticity imaging is inherently ill-posed because the stress distribution inside the domain cannot be measured. Many experimental, theoretical, and numerical studies over the past three decades have tackled this topic, and various methods have been introduced to solve the inverse problem [1,2].

In recent years, methods that employ neural networks with physics-based loss functions to solve inverse problems have become popular [3–7]. In these methods, fully connected feed forward networks estimate mechanical parameters (and stress fields) by taking spatial coordinates as inputs. The outputs are then placed in respective physical equations to construct physics-based loss functions. As static equilibrium equations (or more generally balance of linear momentum) equations in mechanics contain partial derivatives of mechanical stress, physics-informed neural networks (PINN) methods use automatic differentiation to compute these partial derivatives [3,4,6,7] or alternative methods such as convolution kernels to model the equilibrium [5,8].

Fully-connected networks are not the most efficient choice for learning from spatially structured data [9]. This type of data, which includes data acquired from most imaging modalities, is arranged in a way that preserves the spatial relationships between the different data points. While fully connected approaches are highly expressive and powerful in learning complex nonlinear relationships between inputs and outputs, they take a long time to learn complex spatial patterns [7]. In addition, they become increasingly costly to train for deep networks or large datasets. Convolutional neural networks (CNNs), on the other hand, are best suited for tasks that require processing spatially structured data by sharing weights and pooling layers for different regions of the image or volume. These networks seem particularly promising to infer the nonlinear transformation between, say, strain distributions and elasticity parameter fields by satisfying the governing physical equations.

Several studies have already demonstrated the power of physics-informed models with CNNs and UNet structures (encoder-decoder CNN with skip connections between the encoder and decoder paths) in applied mathematics, physics, and engineering applications. These models leverage the trainability of these image-to-image networks as operators on spatially structured input data. Physics-informed UNets have been used as a super-resolution tool conserving equilibrium constraints from low-resolution simulated solid mechanics loadings [10]. Surrogate modeling is another area where CNNs [11], UNets [12] or a combination of multi-task learning and attention UNets [13] have been employed to solve multiple forward problems and generalize to new input information, or aid in learning from sparse training data. UNets have also shown great efficiency in learning directly from physics data when coupled with vision transformers [14]. These examples show the versatility of this type of network in image-to-image tasks in scientific machine learning. To the best of our knowledge, UNets have not been used to directly solve inverse problems in elasticity using only physics constraints.

Elasticity imaging inverse methods need relevant benchmarking examples to evaluate their performance in reconstruction of material parameter fields. These examples often involve circular or elliptical shapes embedded in a uniform background, replicating tumorous tissue behavior [6,15,16]. However, more complex and biologically relevant spatial distributions can demonstrate the robustness of these inverse methods more comprehensively and present them as potential tools for characterization of tissues across multiple scales. Brain tissue is comprised of many tissue subtypes with varying material properties as well as complex geometrical patterns [17–19], rendering it an excellent benchmarking example. Furthermore, reliable mechanical characterization of the brain is crucial in clinical decision making and informing models of extremely important health issues such as traumatic brain injury and surgical planning [20,21].

We present El-UNet, an inversion physics-based neural network model based on the UNet encoder-decoder structure, to solve inverse problems in linear elasticity. Our model solves the material parameter and stress distributions by taking normalized strain distributions as input images and boundary and domain physics information for loss function. We propose several El-UNet implementations, including two with self-adaptive spatial loss weighting methods, and compare how they affect accuracy in space-dependent estimation of isotropic linear elasticity parameters in a heterogeneous 2D example. We also demonstrate how these models perform compared to the fully-connected (dense) PINN implementation under equal circumstances. We show the performance of the models in estimating material parameters on three embedded brain tissue examples with distinct assignment of elastic modulus and Poisson's ratio for white matter, gray matter, and the background. Two examples differ in whether the background region is stiffer or softer than the brain and the other example involves a noisy strain input. These benchmarking examples reveal the robustness of the various tested models against various characterization scenarios.

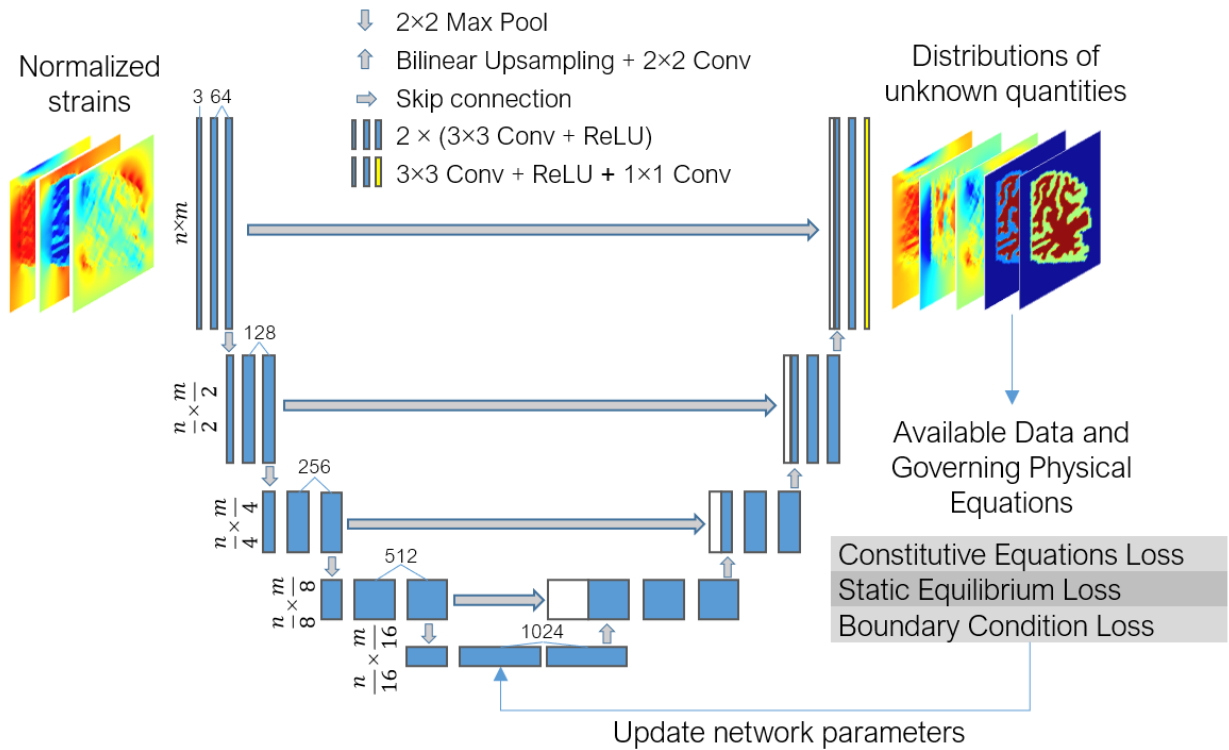

Figure 1. General overview of UNet for inversion in elasticity (El-UNet) implementation. Spatial distributions of strains are fed as three input channels ($\varepsilon_{xx}$, $\varepsilon_{yy}$, and $\varepsilon_{xy}$) to the UNet. The encoder-decoder network is five levels deep, increasing in number of channels from 64 in the shallowest level to 1024 in the deepest level on both the encoder and decoder sides. The final stage has two or five output channels, depending on whether only material parameters or both material parameters and stress terms are outputted. The network outputs enter physical and boundary mean squared error loss equations and the Adam optimizer acts on the sum of the loss functions and updates the network parameters. This loop is repeated until training finishes.

# 2 Methods

## 2.1 Isotropic Formulation

The elasticity equation in index notation is written as:

$$\sigma_{ij} = C_{ijlm}\varepsilon_{lm} \tag{1}$$

where $\sigma$ and $\varepsilon$ are the stress and strain tensors, respectively and $C$ is the stiffness matrix. For isotropic linear elasticity in two dimensions, the equations reduce to

$$\begin{Bmatrix}\sigma_{xx}\\ \sigma_{yy}\\ \sigma_{xy}\end{Bmatrix} = \begin{bmatrix} 2\mu+\lambda & \lambda & 0\\ \lambda & 2\mu+\lambda & 0\\ 0 & 0 & \mu\end{bmatrix}\begin{Bmatrix}\varepsilon_{xx}\\ \varepsilon_{yy}\\ 2\varepsilon_{xy}\end{Bmatrix} \tag{2}$$

where $\lambda$ and $\mu$ are the Lamé parameters [22]. Elastic modulus and Poisson's ratio for a plane strain problem can be derived from the Lamé parameters using:

$$\begin{aligned} E &= \frac{\mu(3\lambda+2\mu)}{\lambda+\mu}\\ \nu &= \frac{\lambda}{2(\lambda+\mu)}. \end{aligned} \tag{3}$$

For plane stress assumptions, the following conversion should be applied when solving the inverse problem:

$$E_{plane\ stress} = \frac{E_{plane\ strain}}{1 - \nu_{plane\ strain}^2}, \qquad \nu_{plane\ stress} = \frac{\nu_{plane\ strain}}{1 - \nu_{plane\ strain}}. \tag{4}$$

The static equilibrium equations after neglecting body forces in the system reduce to

$$\begin{aligned}\frac{\partial \sigma_{xx}}{\partial x} + \frac{\partial \sigma_{xy}}{\partial y} &= 0\\ \frac{\partial \sigma_{xy}}{\partial x} + \frac{\partial \sigma_{yy}}{\partial y} &= 0.\end{aligned} \tag{5}$$

We implement a dimensionless variation of the above equations in the inversion algorithm and use mean dimensions of the geometry ($l_0$) and maximum normal stress on the traction boundary ($\sigma_0$) as reference characteristic scales. Therefore Equations 2 and 5 can be written as:

$$S_{xx} = (2M + \Lambda)\varepsilon_{xx} + \Lambda\varepsilon_{yy}$$
$$S_{yy} = (2M + \Lambda)\varepsilon_{yy} + \Lambda\varepsilon_{xx} \quad (6)$$
$$S_{xy} = 2M\varepsilon_{xy}$$

$$\frac{\partial S_{xx}}{\partial X} + \frac{\partial S_{xy}}{\partial Y} = 0$$
$$\frac{\partial S_{xy}}{\partial X} + \frac{\partial S_{yy}}{\partial Y} = 0. \quad (7)$$

where the upper-case letters denote dimensionless values. In-depth details regarding the dimensionless approach can be found in our previous publication [7].

## 2.2   Finite Element Simulation

We performed a finite element simulation of brain slice under tensile loading, as detailed in our previous work [7]. In brief, we collected a T1-weighted image of a 28-year-old male subject in a 3.0 Tesla MRI Scanner (Skyra, Siemens Healthcare, Germany) and a 32-channel head coil (human subject imaging approved by University of Arizona Institutional Review Board, February 2020). Next, we picked a coronal slice near the posterior side of the brain, segmented gray matter and white matter using a threshold, and developed a finite element model of the brain slice in ANSYS Workbench (Ansys, Inc., PA, USA), embedded in a rectangular hydrogel background. Finally, we loaded the entire specimen from the top side with uniform normal stress in the vertical direction, chosen to result in nominal axial strains not larger than 5% anywhere in the domain while keeping the bottom side a frictionless boundary. Here, the background material was chosen to be softer than the brain slice (1kPa background vs 1.5kPa/2kPa gray matter/white matter) as presented in Table 1. As a secondary example, we also performed a simulation with stiffer background material (5 kPa). We used data from the finite element simulation as input data (strains and stress boundary conditions) as well as ground truth to compare the model output against (full field material parameter and stress distributions). As a separate example, we also added 10% Gaussian noise to the strain data from the soft background example to study the robustness of the inverse models against noisy strain images. These examples comprised a variety of conditions that allowed us to evaluate the performance of the models in various scenarios (Table 1). As we reported in our previous work [7], the brain tissue geometry can be considered a complex yet biologically relevant benchmarking example. The models that accurately resolve the distribution of patterns for this example are expected to perform equally well or better for simpler heterogeneous patterns and inclusions in elasticity imaging. For the dense PINN runs, we picked training collocation points uniformly from the unstructured mesh [7], whereas for the UNet runs, we used the triangulation-based natural neighbor interpolation in MATLAB (MathWorks, MA, USA) to construct structured isotropic distributions from the unstructured mesh. The domain data prepared for the dense PINN method had 14200 collocation points while the image dimensions for the UNet model was 142*100 resulting in 14200 pixels. Therefore, both network variations had the same data size. Figure 2 shows the strain distribution patterns and material parameter distributions from each example.

**Table 1**. Assigned material properties for finite-element modeling of loaded specimens.

| Material | Elasticity Parameters | |
| --- | --- | --- |
| | $E$ (kPa) | $\nu$ |
| White Matter | 2 | 0.35 |
| Gray Matter | 1.5 | 0.4 |
| Background | 1 (soft background example) <br> 5 (stiff background example) | 0.45 |

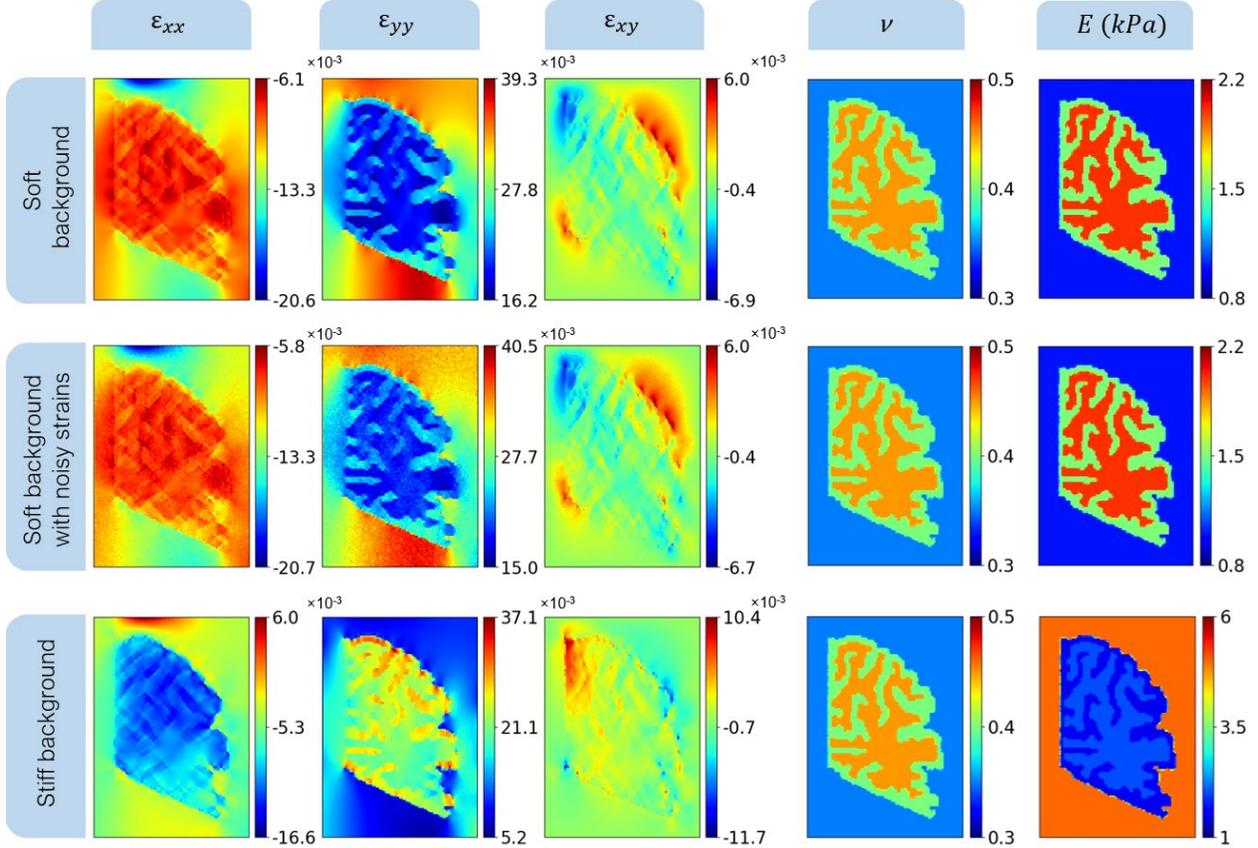

Figure 2. Breakdown of finite element-derived strain fields and parameter distributions used as input and validation data for the inverse models, respectively.

## 2.3 El-UNet Implementation

We developed El-UNet, an encoder-decoder structure based on the original UNet architecture [23], to solve the inverse problem in quasi-static elasticity imaging (Figure 1). In brief, compared to the original work, we removed bias parameters and used batch normalization in the 3×3 double-convolution sections, and upsampling followed by a 2×2 convolution in the upward path of the network instead of transposed convolutions. Each convolution layer was followed by ReLU activation function to introduce non-linearity except for the last layer, which had a linear output. All the convolutions had a stride of one, whereas the pooling layers had a stride of 2. A padding of 1 was used to maintain dimensions after convolutions. We also used resampling steps in the upward path in case the output of the double convolutions had dimensions not matching the skip connection image, which would occur to odd image dimensions due to pooling in the downward path. We used the original number of channels for the double convolutions, i.e., 64, 128, 256, 512, and 1024 channels, respectively, moving in the downward path of the UNet and the reverse trend for the upward path. The network takes in a 3-channel input, each channel containing the normalized spatial distribution of a strain tensor term (two normal and one shear strain distributions) and estimates either dimensionless Lamé parameters only (P El-UNet, Figure 3A) or Lamé parameters and stress distributions together (PS El-UNet, Figure 3B).

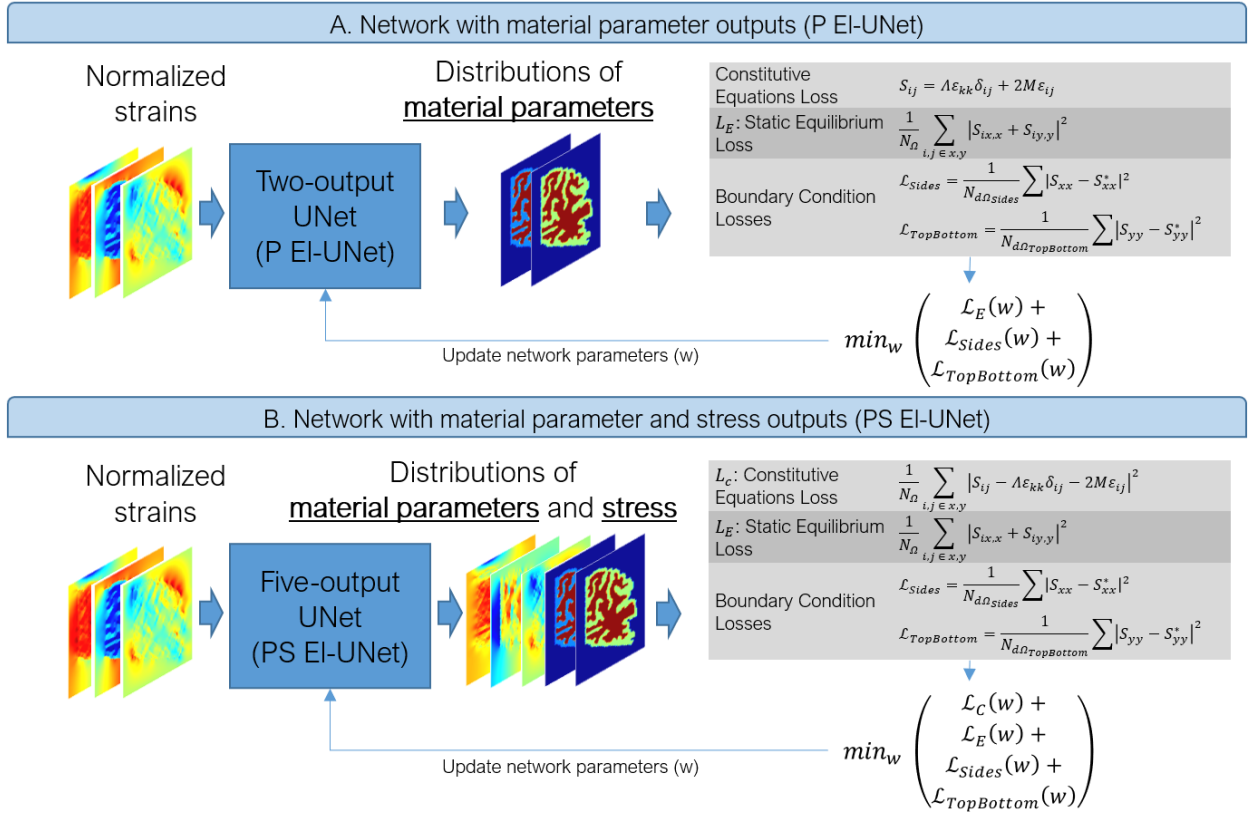

Figure 3. Breakdown of the two main UNet setups used in this study in terms of network output. The main difference between the two is the output channels.

For P El-UNet, the algorithm uses the isotropic linear elasticity constitutive equations to compute stress terms across the domain using the estimated Lamé parameters and given strains. It then computes the mean squared error (MSE) loss values for static equilibrium in two directions and normal stress on the boundaries. The partial derivatives in the static equilibrium equations are approximated as finite central difference inside the domain and forward/backward difference on the boundaries. Following the dimensionless approach, the spacing for the central difference approximation is computed as:

$$\Delta x = length_x/(l_0(N_x - 1))$$
$$\Delta y = length_y/(l_0(N_y - 1))$$

(8)

where $N_x$ and $N_y$ are the number of pixels in x and y, respectively. For PS El-UNet, the algorithm uses a mean-squared error loss function to balance the stress distribution directly estimated by the network and the one computed by plugging output Lamé parameters and given strains in the constitutive equations. The remaining stages of the five-output implementation are like P El-UNet.

## 2.4 Self-adaptive Spatial Loss Weighting

We experimented with two self-adaptive loss weighting methods with the goal of speeding up convergence to accurate parameter distributions and better resolving the complex patterns in the images. We implemented these methods on the PS El-UNet configuration and, thus, named them PS El-UNet W1 and PS El-UNet W2 (Figure 4). In the PS El-UNet W1 configuration, we created three types of trainable weight fields, each with values initialized at 1. These were defined as self-adaptive spatial weights for constitutive equations ($\psi_C$), static equilibrium ($\psi_E$) and boundaries ($\psi_{Sides}$ and $\psi_{TopBottom}$). These spatial weights were multiplied in an element-wise manner by the left-hand side and right-hand side of their corresponding MSE losses and updated in the optimizer along with the network weights in a min-max approach as outlined in Figure 4A. The PS El-UNet W2 had the same setup as W1 except that it did not have the static equilibrium spatial weighting, $\psi_E$ (Figure 4B). A similar strategy was previously used for fully-connected PINNs when solving the forward problem in non-Fourier heat conduction and had shown better convergence compared to the PINN model with no adaptive weighting with equal training epochs [24]. Compared to that study, we use two variations of this method for the inverse elasticity problem and compare their performances with non-adaptive El-UNet in equal computation timeframes to assess the potential accuracy gain under similar computational cost circumstances.

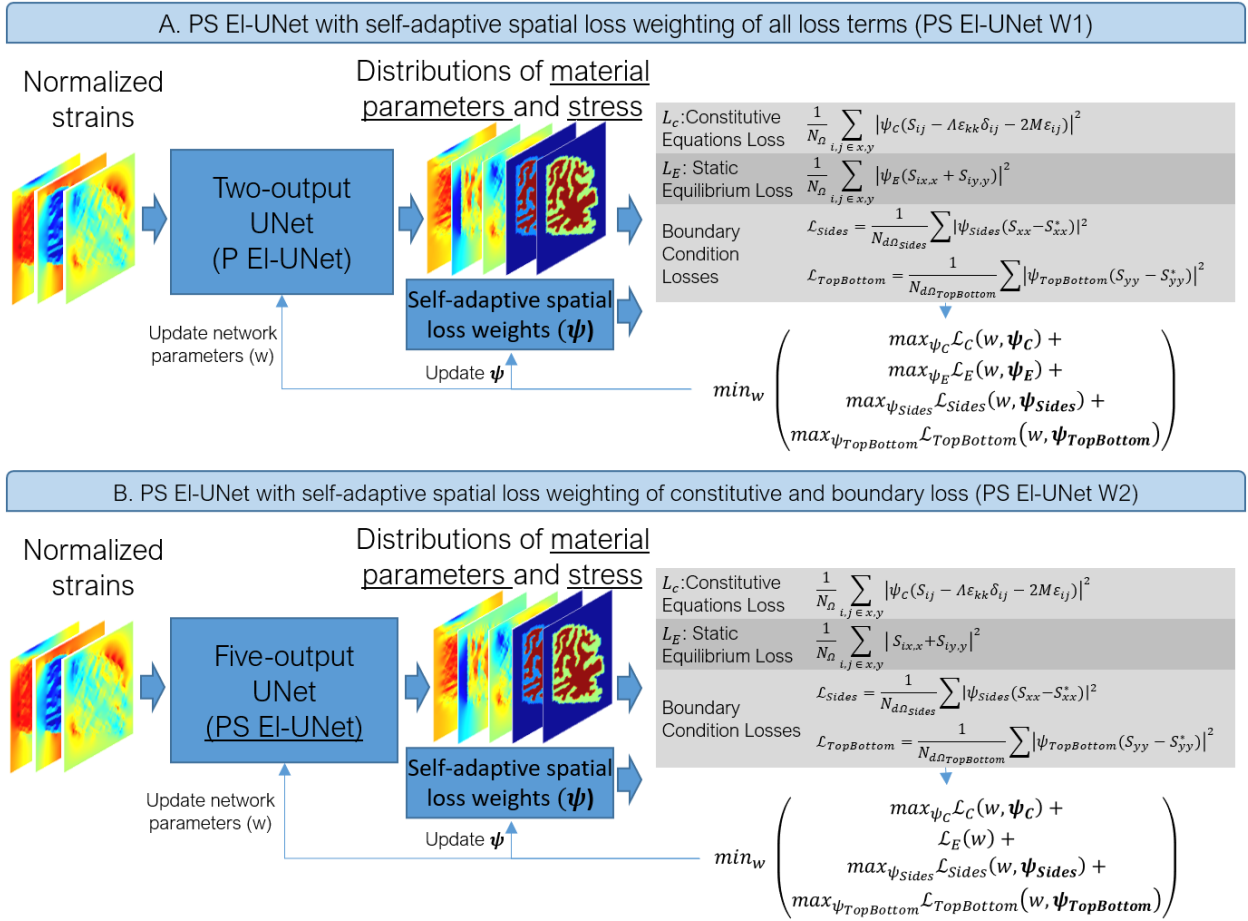

Figure 4. Breakdown of the two self-adaptive spatial weighting approaches. The PS El-UNet W1 configuration has self-adaptive spatial weights for all the loss terms. PS El-UNet W2 is similar except that the static equilibrium loss is not weighted.

## 2.5 Dense PINN

We compared the proposed models with our previously published fully-connected physics-informed neural network implementation for the same task to demonstrate the improvements achieved by the current models. For this purpose, we constructed the networks and the training pipeline as described in our previous study [7]. We used the same resolution of input and boundary condition data as the UNet models to keep the training procedure exactly similar between them except for the model used.

## 2.6 Implementation and Computation Details

We wrote the codes for the UNet and PINN implementations in PyTorch v1.13.1. For all the models, we used the Adam optimizer with a learning rate of 0.001 with no decay settings to minimize the loss value and trained each model for 30 minutes on Nvidia P100 GPUs. Due to the oscillatory nature of loss evolution through the training process, we performed each run ten times, plotted the average output for visualizations, and reported means and standard deviations of quantified errors where applicable. Because the network state in each run was initiated randomly, the number of epochs performed during each run in the equal time given had a small variance. Therefore, we plotted loss and error vs. epoch number trends up to the minimum epoch number that all models reached for that specific example and model configuration. For the spatially weighted runs, while the optimizer acted on the weighted loss of the model, here, we report the loss associated with the physical equations in their non-weighted state. This reporting approach allows us to compare weighted with non-weighted models in terms of the minimization of the physics-associated loss values. Table 2 provides a short description of all the models investigated in this study.

**Table 2**. Description of the physics-informed inversion models under study.

| Model | Description |
| --- | --- |
| Dense PINN | Two multilayered perceptrons, outputting parameter and stress distributions (from [7]) |
| P El-UNet | UNet architecture with material parameter distributions as outputs and normalized strain images as input channels |
| PS El-UNet | UNet architecture with material parameter and stress distributions as outputs and normalized strain images as input channels |
| PS El-UNet W1 | PS El-UNet configuration with self-adaptive spatial loss weighting for all loss terms |
| PS El-UNet W2 | PS El-UNet configuration with self-adaptive spatial loss weighting for constitutive equations and boundary conditions |

# 3 Results

## 3.1 Visual Depiction of Estimated Fields

The UNet results were generally more accurately resolved compared to Dense PINN (Figures 5-7). Starting with the example with soft background (Figure 5), the two-output implementation showed visible artifacts, especially for the $v$ estimation. The Five-output implementation did not have these artifacts but looked less accurate in terms of overall discovered patterns. The two spatially weighted implementations were the closest estimation to ground truth with little artifacts.

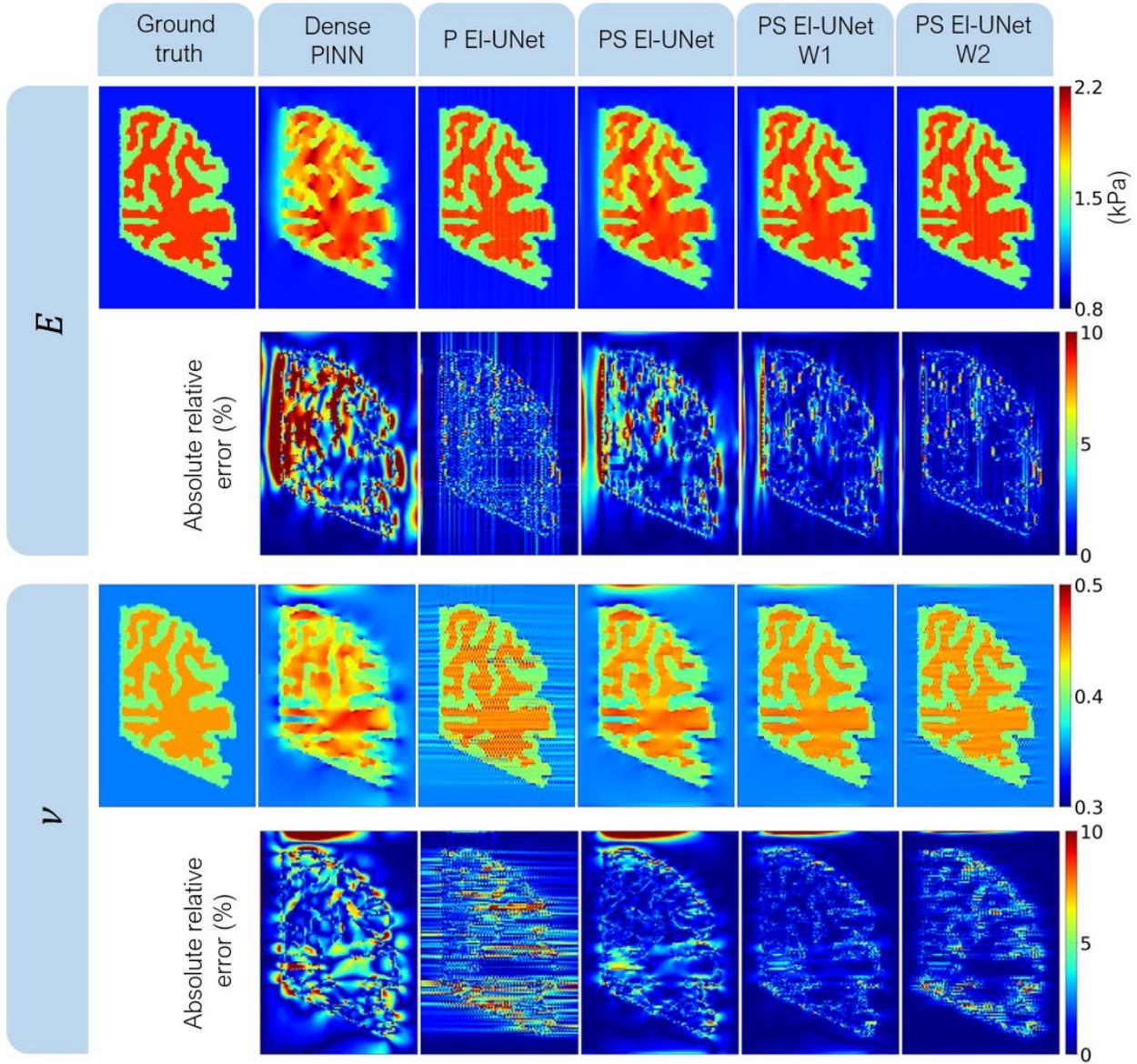

Figure 5. Estimation and absolute relative error maps from the various physics-informed models under study for the soft background example. Qualitative evaluation of estimated maps reveals improved estimation of El-UNet models compared to Dense PINN. The weighted Unet models, namely W1 and W2 show the best results for both $E$ and $v$.

Regarding the inverse run with noisy strain data, the UNet models resolved a grainy reconstruction of the unknown parameters (Figure 6). Here, the PINN model produced less grainy outputs but did not resolve the pattern as thoroughly as the UNet models.

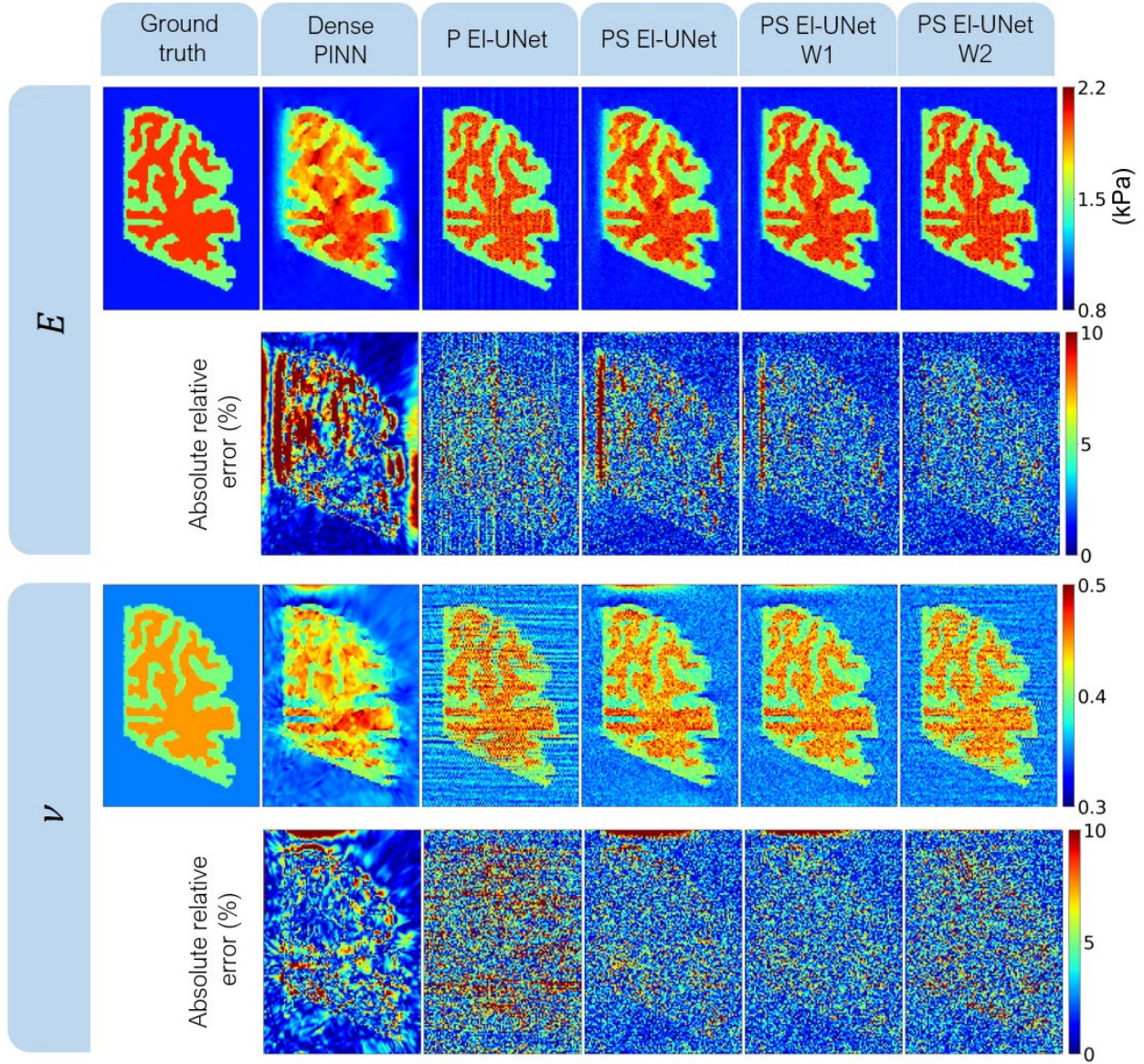

Figure 6. Estimation and absolute relative error maps from the various physics-informed models under study for the soft background example with noisy strain inputs. The dense PINN model provides less accurate and less grainy reconstruction than unweighted UNet models. The weighted and unweighted PS El-UNet models show more robustness against noise.

The stiff background example had the worst PINN and P El-UNet reconstruction of unknown parameters among the studied examples (Figure 7). The $v$ transition between the background and the gray matter was specifically poorly reconstructed. PS El-UNet W1 worked better than other models for the same example.

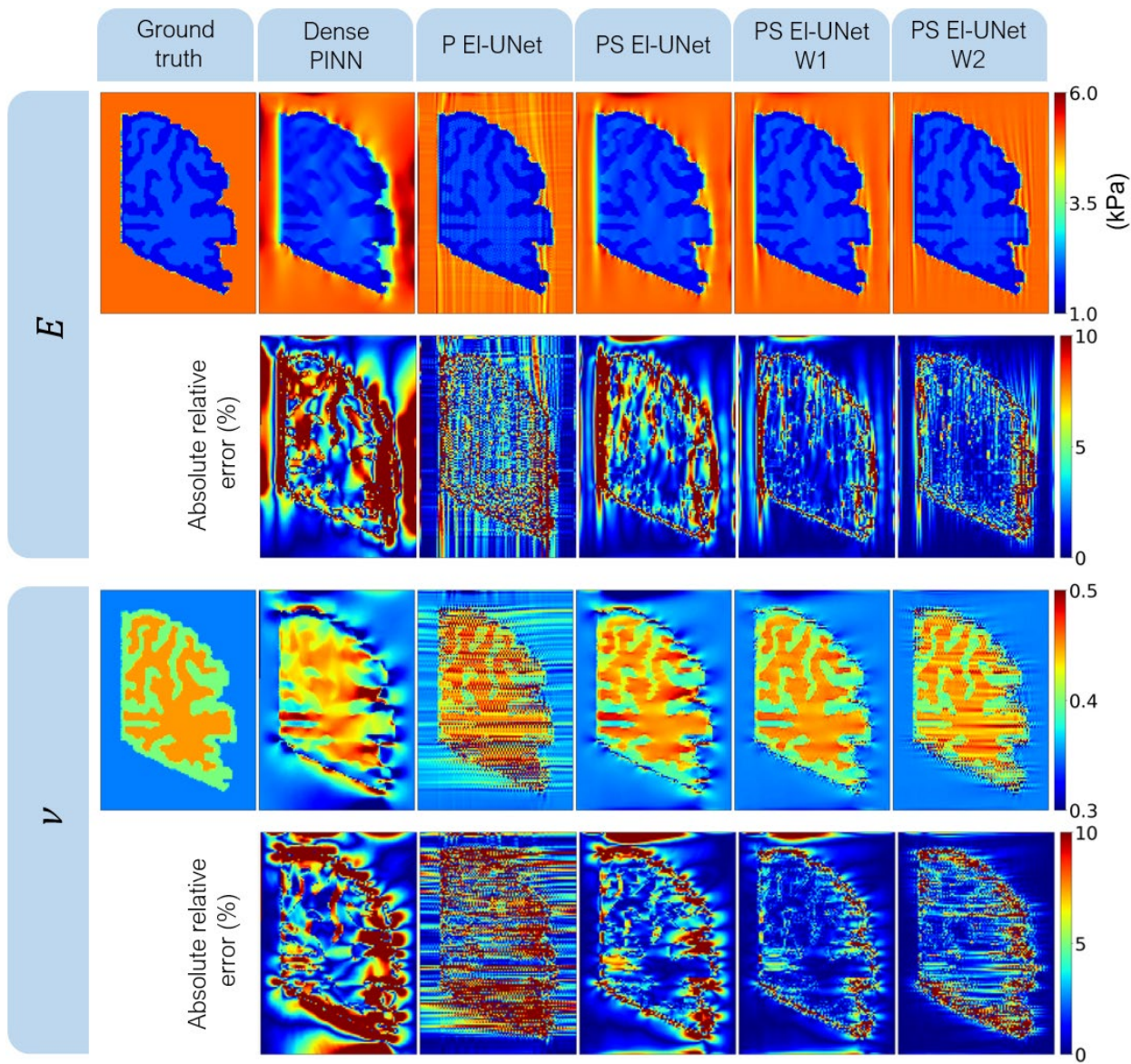

Figure 7. Estimation and absolute relative error maps from the various physics-informed models under study for the stiff background example. The transition zone between the background and gray matter had higher reconstruction errors ompared to soft background examples. PS El-UNet W1 shows the best estimation in terms of capturing the complex pattern while minimizing reconstruction artifacts.

## 3.2 Quantified Loss and Estimation Errors

To analyze model performance more objectively, we compared the evolution of total loss and mean estimation errors between the models (Figure 8). Comparing the two-output network (P El-UNet) and the five-output network (PS El-UNet) with PINN showed that while both models outperform the Dense PINN in E estimation accuracy, PS El-UNet has better $v$ estimation. Between the self-adaptive weighted configurations, PS El-UNet W1 showed estimation errors that either kept decreasing or almost plateaued at low values, while W2 showed a reversal for the $v$ estimation error in later epochs, especially visible with noisy strain and stiff background examples.

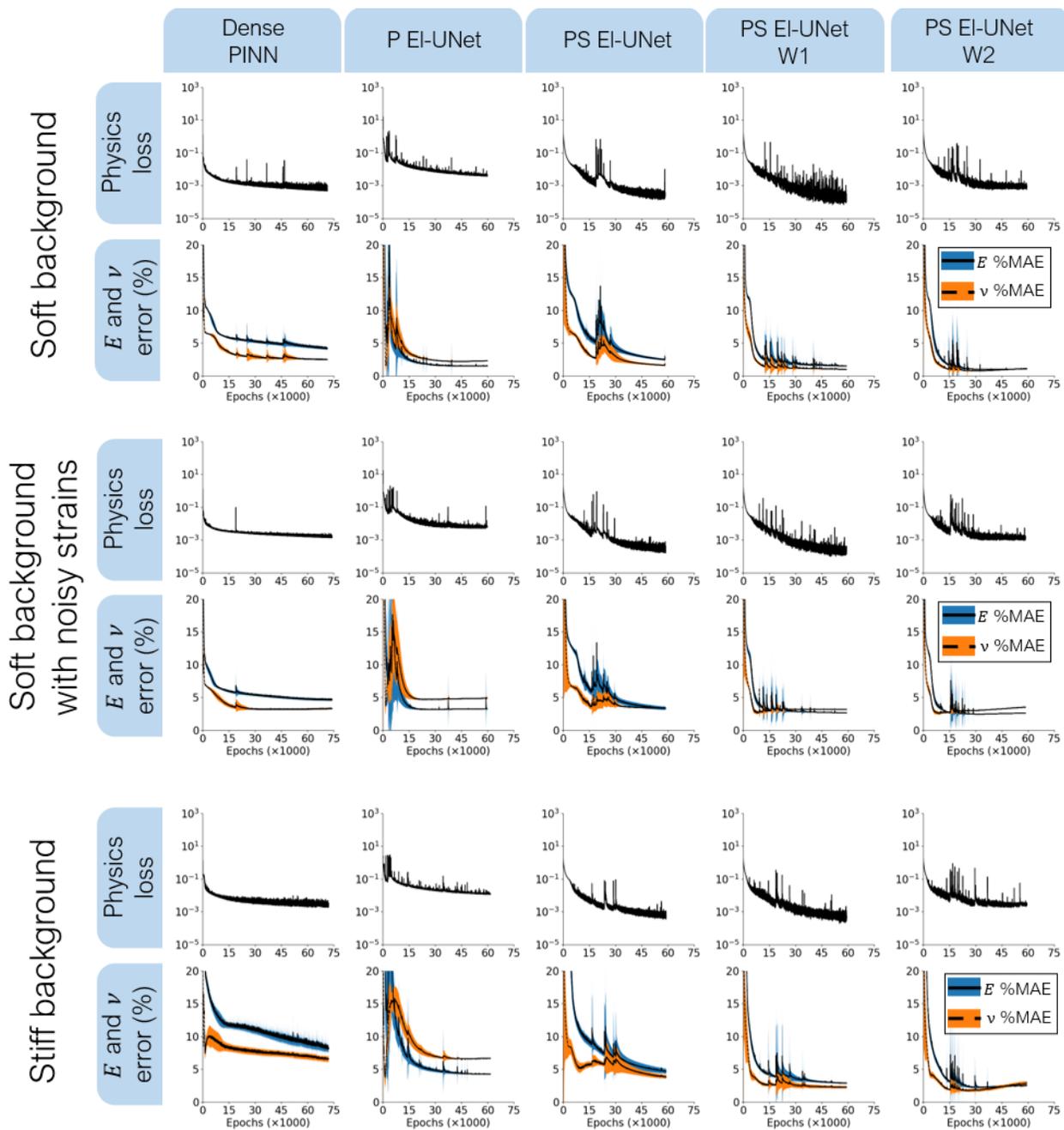

Figure 8. Total training loss and mean absolute relative errors (solid lines) along with standard deviation (shades) associated with $E$ and $v$ estimation. PS El-UNet W1 has the most reliable loss and error evolution trend across epochs.

## 3.3 Self-adaptive Spatial Loss Weight Distributions

The final spatial distribution of the self-adaptive spatial loss weights from the different examples allows us to interpret the improved performance of the weighted models (Figure 9). The regions where the loss values were larger reflect where the model learned to give more weight to the corresponding loss term, i.e., constitutive equations ($\psi_C$), static equilibrium ($\psi_E$) and boundaries ($\psi_{Sides}$ and $\psi_{TopBottom}$).

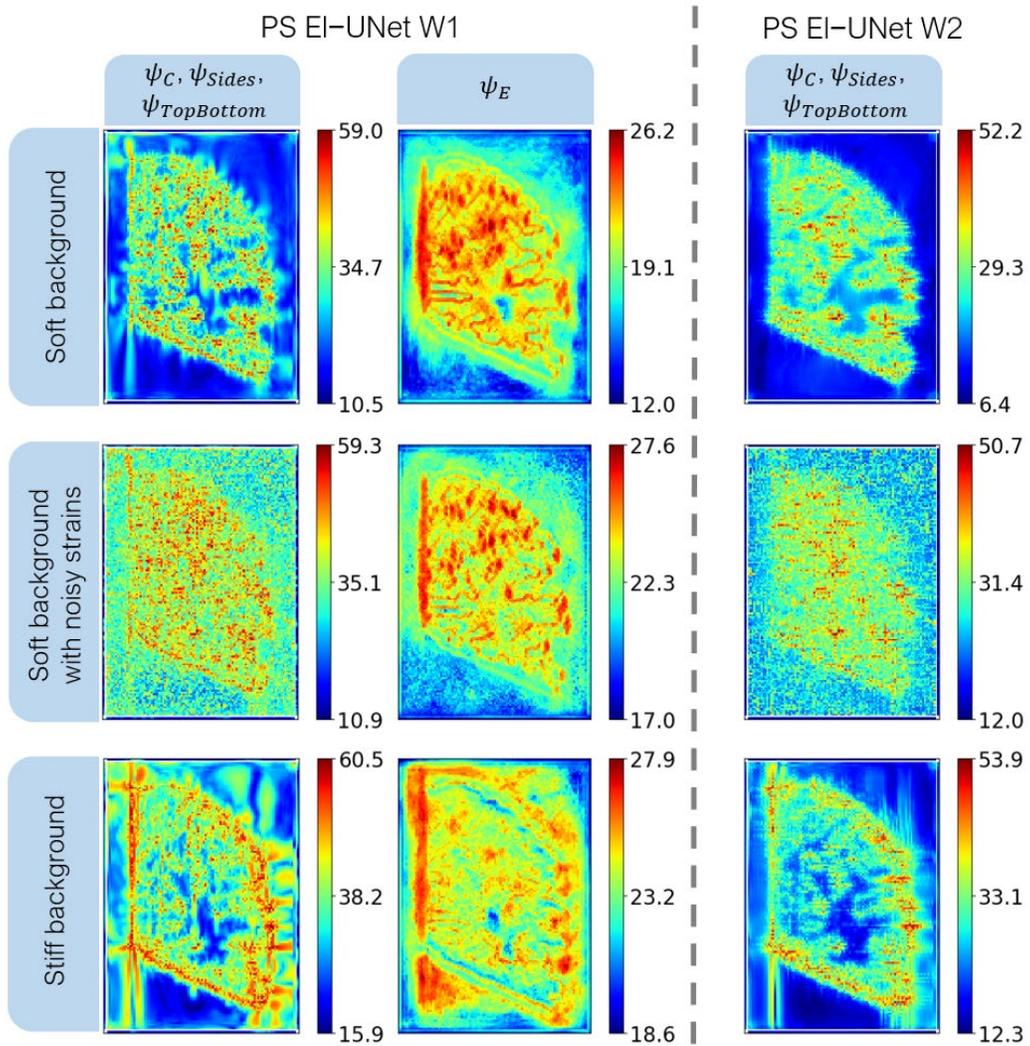

Figure 9. The final state of learned self-adaptive spatial loss weights from the different examples and the two weighted models under study. constitutive equations weight: $\psi_C$, static equilibrium weight: $\psi_E$, and boundary weights: $\psi_{Sides}$ and $\psi_{TopBottom}$. Boundary weights are plotted with a pixel offset around the $\psi_C$ map. The models learn to penalize themselves more in the regions of the image where violations of the physical constraints are highest during training.

# 4 Discussion

We introduced UNet-based models for inverse reconstruction of material properties from strain fields and boundary conditions in elasticity imaging, collectively named El-UNet. This paper focused on different variations of these models for 2D plane stress examples and compared their performance with one another and our previous model, which used densely connected physics-informed neural networks. The results visibly showed improved reconstruction by the UNet-based models, with the spatially weighted models showing the best performance. The weighted models were the fastest and achieved lowest estimation errors among the different tested models under similar computational time circumstances. The final weight distribution indicated the areas the model learns to penalize itself more while training. Tracking the error decay patterns across epochs revealed that the weighted models reach the lowest estimation errors and effectively discover the complex patterns much faster than the alternatives, making them ideal for larger datasets such as volumetric elasticity imaging.

The encoder-decoder structure and convolutional nature of the UNet clearly showed advantages over fully-connected implementation of PINN. The convolution kernels share weights for the different parts of the image and recover patterns better, making them ideal for spatially structured data such as images and volumes. Comparing PINN with P El-UNet and PS El-UNet clearly shows El-UNet's superior performance in resolving accurate distribution of unknown parameters in the relatively short estimation time shared between all models. These improvements are evident in successful reconstruction of sharp gradients between different regions of the image. Moreover, trends of error decay through the course of training for the different models show that while PINN reaches low errors faster than the unweighted El-UNet models, the errors almost plateau and the later updates only slowly decrease the estimation error of the unknown parameters, while El-UNet reaches lower estimation errors for unknown parameters.

Determining better performance between the models in terms of output type (P El-UNet vs PS El-UNet) came down to existence of artifacts in the reconstructed fields. In P El-UNet, the model only estimates the unknown parameter distributions and plugs those values, along with strains, into the constitutive equations to compute stress distributions. The partial derivatives of these stress values are then used to satisfy the static equilibrium equations. The stress in this configuration becomes directly correlated with material parameters and the finite difference approximation amplifies the error that exists in the output of the network. In addition, for the noisy strain case, the noise directly affects the computed stress, and enforcing static equilibrium equation is affected by the first derivatives of these noisy stress fields. Conversely, in the PS El-UNet, the model becomes better regularized by enforcing the constitutive equations as soft constraints. In other words, the MSE loss of the equilibrium equations has stress terms that are independent outputs of the network, themselves separately balanced by the constitutive equations' MSE loss in a soft manner. We observed the implications of this network design choice by comparing PS El-UNet and P El-UNet outputs for the various examples in this study.

We improved the convergence of the UNet-based models with the introduction of self-adaptive spatial loss weights with two proposed weighting schemes. The two models differed in whether they were weighted for all their loss terms or only constitutive equations and boundary conditions. The results clearly showed that both weighted implementations visibly led to better reconstruction than the unweighted approaches. Tracking loss and mean estimation error values for the unknown parameters across epochs revealed that PS El-UNet W2 has a reversal behavior of $v$ mean estimation error in the noisy and stiff background examples. We speculate that when the static equilibrium loss is unweighted, the balance between the static equilibrium loss and constitutive equations loss tips too much over to the latter leading to reconstructions with artifacts. The final distribution of spatial loss weights shows the increased intensities corresponding to regions where the model learned to penalize itself more. Comparing these distributions with the estimation fields and associated error maps reveals the high-intensity weight regions overlap with high estimation error regions of the unweighted PS El-UNet model. Previous work on physics-informed neural networks has shown the imbalance existing between the multi-objective loss terms resulting in poor convergence. However, the self-adaptive loss term weighting presented in these studies requires additional backpropagation for the optimizer update, does not impose spatial weighting, and has only been tested in fully-connected networks [25,26]. Another study on using PINN in linear elastic micromechanics proposed a dynamic weighting approach that increased the density of collocation points in the regions of the domain with high losses, effectively increasing contribution of the errors associated with those points to the loss function [27]. In the self-adaptive spatial loss weighting method presented in the current study, the additional weights do not belong to any extra deep network as they are merely trainable parameters. Therefore, the optimizer updates do not require backpropagation through an entire network for each update. The size of the input space does not change either during training as a result of the weight updates. This configuration ensures that the weighted models perform with almost the same speed as the unweighted PS El-UNet model, as evidenced by comparing the number of finished epochs in the same duration between these models. This is an important implication of this approach because, at similar computational costs, we can recover more accurate results without a priori knowledge of the problem at hand and the material distributions.

It is worthwhile to mention a few limitations of El-UNet. The current model works with isotropic spatially structured data. While the examples covered in this work all had isotropic resolutions, anisotropic resolutions e.g., from ultrasound and magnetic resonance images, can be integrated into the model by appropriate unequal differentiation intervals in the finite difference computation stage. Although elastography images are usually stored as rectangular structured grids, a method to map non-rectangular domains to rectangular ones to benefit from convolutional neural networks has been reported in the literature [28]. We used the simplest approximation for partial derivatives in the static equilibrium equations, which was prone to error amplification in some variations of our model. Alternative implementations have also been proposed to pose the static equilibrium equations in the form of convolutional layers that integrate well with the network and could potentially avoid the error-accumulation drawbacks of finite-difference approximation [5,8,11]. Finally, estimation for more complex problems or material models can be improved by more advanced image-to-image networks.

The main ideas presented in our study, namely using UNet-based models for physics-informed inversion and spatial loss weighting, are the first steps to scale to 3D estimations and other material models such as multi-parameter orthotropic elasticity or hyperelasticity, both of which are relevant models in biological tissues.

# 5 Acknowledgments

**Funding**: This work was supported by the National Institutes of Health (NIH) National Institute of Biomedical Imaging and Bioengineering (NIBIB) Trailblazer award number R21EB032187.

# 6 Competing Interests Statement

The authors declare no conflict of interest.

# 7 References

[1]     M.M. Doyley, Model-based elastography: A survey of approaches to the inverse elasticity problem, Phys. Med. Biol. 57 (2012).

[2]     M.M. Doyley, K.J. Parker, Elastography: general principles and clincial applications, Ultrasound Clin. 9 (2014) 1.

[3]     E. Zhang, M. Yin, G.E. Karniadakis, Physics-informed neural networks for nonhomogeneous material identification in elasticity imaging, ArXiv Prepr. ArXiv2009.04525. (2020).

[4]     E. Haghighat, M. Raissi, A. Moure, H. Gomez, R. Juanes, A physics-informed deep learning framework for inversion and surrogate modeling in solid mechanics, Comput. Methods Appl. Mech. Eng. 379 (2021) 113741.

[5]     C.T. Chen, G.X. Gu, Learning hidden elasticity with deep neural networks, Proc. Natl. Acad. Sci. U. S. A. 118 (2021).

[6] E. Zhang, M. Dao, G.E. Karniadakis, S. Suresh, Analyses of internal structures and defects in materials using physics-informed neural networks, Sci. Adv. 8 (2022).

[7] A. Kamali, M. Sarabian, K. Laksari, Elasticity imaging using physics-informed neural networks: Spatial discovery of elastic modulus and Poisson's ratio, Acta Biomater. 155 (2023) 400–409.

[8] C.T. Chen, G.X. Gu, Physics-Informed Deep-Learning For Elasticity: Forward, Inverse, and Mixed Problems, Adv. Sci. 2300439 (2023) 1–11.

[9] L. Yann, B. Yoshua, Convolutional Networks for Images, Speech, and Time-Series, Handb. Brain Theory Neural Networks. (1995) 255–258.

[10] K. Yonekura, K. Maruoka, K. Tyou, K. Suzuki, Super-resolving 2D stress tensor field conserving equilibrium constraints using physics-informed U-Net, Finite Elem. Anal. Des. 213 (2023) 103852.

[11] Y. Zhu, N. Zabaras, P.S. Koutsourelakis, P. Perdikaris, Physics-constrained deep learning for high-dimensional surrogate modeling and uncertainty quantification without labeled data, J. Comput. Phys. 394 (2019) 56–81.

[12] X. Zhao, Z. Gong, Y. Zhang, W. Yao, X. Chen, Physics-informed convolutional neural networks for temperature field prediction of heat source layout without labeled data, Eng. Appl. Artif. Intell. 117 (2023) 105516.

[13] Z. Cao, W. Yao, W. Peng, X. Zhang, K. Bao, Physics-Informed MTA-UNet: Prediction of Thermal Stress and Thermal Deformation of Satellites, Aerospace. 9 (2022) 1–16.

[14] O. Ovadia, A. Kahana, P. Stinis, E. Turkel, G.E. Karniadakis, ViTO: Vision Transformer-Operator, (2023).

[15] P.E. Barbone, A.A. Oberai, Elastic modulus imaging: Some exact solutions of the compressible elastography inverse problem, Phys. Med. Biol. 52 (2007) 1577–1593.

[16]   M.T. Islam, S. Tang, C. Liverani, S. Saha, E. Tasciotti, R. Righetti, Non-invasive imaging of Young's modulus and Poisson's ratio in cancers in vivo, Sci. Rep. 10 (2020) 1–12.

[17]   S. Chatelin, A. Constantinesco, R. Willinger, Fifty years of brain tissue mechanical testing: From in vitro to in vivo investigations, Biorheology. 47 (2010) 255–276.

[18]   K. Laksari, M. Shafieian, K. Darvish, Constitutive model for brain tissue under finite compression, J. Biomech. 45 (2012) 642–646.

[19]   S. Budday, T.C. Ovaert, G.A. Holzapfel, P. Steinmann, E. Kuhl, Fifty Shades of Brain: A Review on the Mechanical Testing and Modeling of Brain Tissue, Springer Netherlands, 2019.

[20]   J.S. Giudice, W. Zeng, T. Wu, A. Alshareef, D.F. Shedd, M.B. Panzer, An Analytical Review of the Numerical Methods used for Finite Element Modeling of Traumatic Brain Injury, Ann. Biomed. Eng. (2018).

[21]   K. Miller, G.R. Joldes, G. Bourantas, S.K. Warfield, D.E. Hyde, R. Kikinis, A. Wittek, Biomechanical modeling and computer simulation of the brain during neurosurgery, Int. j. Numer. Method. Biomed. Eng. 35 (2019) 1–24.

[22]   M.H. Sadd, Elasticity: theory, applications, and numerics, Academic Press, 2009.

[23]   O. Ronneberger, P. Fischer, T. Brox, U-net: Convolutional networks for biomedical image segmentation, in: Int. Conf. Med. Image Comput. Comput. Interv., Springer, 2015: pp. 234–241.

[24]   S. Shi, D. Liu, Z. Zhao, Non-Fourier Heat Conduction based on Self-Adaptive Weight Physics-Informed Neural Networks, Chinese Control Conf. CCC. 2021-July (2021) 8451–8456.

[25]   S. Wang, Y. Teng, P. Perdikaris, Understanding and mitigating gradient flow pathologies in physics-informed neural networks, SIAM J. Sci. Comput. 43 (2021) A3055–A3081.

[26]   S. Wang, X. Yu, P. Perdikaris, When and why PINNs fail to train: A neural tangent kernel

perspective, J. Comput. Phys. 449 (2022) 110768.

[27]  A. Henkes, H. Wessels, R. Mahnken, Physics informed neural networks for continuum micromechanics, Comput. Methods Appl. Mech. Eng. 393 (2022) 114790.

[28]  H. Gao, L. Sun, J.X. Wang, PhyGeoNet: Physics-informed geometry-adaptive convolutional neural networks for solving parameterized steady-state PDEs on irregular domain, J. Comput. Phys. 428 (2021) 110079.